# Batch Transformer Architecture: Case of Synthetic Image Generation for Emotion Expression Facial Recognition

Stanislav Selitskiy
PhD student
University of Bedfordshire, School of Computer Science and Technology
University Square, LU1 3JU, Luton, UK
+1- 678-595-4674
stanislav.selitskiy@study.beds.ac.uk

**Abstract**

A novel Transformer variation architecture is proposed in the implicit sparse style. Unlike "traditional" Transformers, instead of attention to sequential or batch entities in their entirety of whole dimensionality, in the proposed Batch Transformers, attention to the "important" dimensions (primary components) is implemented. In such a way, the "important" dimensions or feature selection allows for a significant reduction of the bottleneck size in the encoder-decoder ANN architectures. The proposed architecture is tested on the synthetic image generation for the face recognition task in the case of the makeup and occlusion data set, allowing for increased variability of the limited original data set.

**Keywords**: attention, Transformer, synthetic image, data augmentation, emotion recognition

## 1. Introduction

Face recognition (FR) is a topic of intensive research on the intersection of pattern recognition and computer vision. Because of its naturally innate nature for humans and noninvasive nature, FR earned its leading place in numerous practical applications in information security, surveillance systems, law enforcement, access control and smart cards. For a decade already, with the pioneering AlexNet architecture, Deep Learning (DL) models for FR have surpassed human accuracy in the controlled environment. However, in real-life settings, especially under conditions that DL models have not seen during training, such as the so-called Out-of-Distribution (OOD) condition, even advanced models suffer from uncertainty and failures.

However, in the case of Facial Expression Recognition (FER), advanced DL models of various architectures (from simple Engineering Features to Convolutional and Transformer ANNs) demonstrate much lower accuracy metrics than for FR and are prone to catastrophic failures, i.e. errors with high confidence.

Therefore, the idea that the whole spectrum of emotion expressions can be reduced to six basic facial feature complexes (Ekman and Friesen 1971) was challenged in the sense that human emotion recognition is context-based, fuzzy, and has many more components than the "basic six". The same facial feature components and their complexes may be interpreted differently depending on the situational context (Cacioppo et al. 2000, Gross and Canteras 2012).

Yet another concern is that the majority of the FER data sets are staged, either with the help of professional actors or plain amateurs, and their expressions are not natural, missing important micro-expressions generated by emotion states in the real, naturalistic situations.

The study aims to improve the accuracy of the DL models by expanding the training set with synthetically generated images. The scope is limited by the study of the auto-encoder ANN architecture based on a Transformer-type attention mechanism for synthetic image generation. The hypothesis we will investigate is: whether the scaling functionality of the Transformers' dot-product attention mechanism, applied to images of subjects expressing different varieties of emotion expressions and microexpressions, generate synthetic images that would cover missing variants in the training data set, improving the DL model performance on the test data set?

The intermediate objectives are to modify and improve the problematic aspects of the dot-product attention mechanism to make the encoder-decoder model viable on the limited hardware resources.

The contribution is organized in the following way: current Section 1 overviews the state and problems associated with Machine Learning algorithms employed in Face and Facial Expression Recognition and highlights aims, research questions, and the scope of the research. Section 2 reviews the current literature in the area of the proposed scope and research questions, Section 3 - describes the methodology of the proposed solution for the research questions; Section 4 - data sets which were used in computational experiments. Section 5 lists hardware parameters and model configurable parameters. Section 6 shows the results of the experiments in diagram and table form; Section 7 draws conclusions, briefly reiterates study limitations, and outlines future research directions.

## 2. Literature Review

### 2.1 Face and Facial Expression Recognition

The most obvious and non-invasive techniques for person recognition are based on biometric measurements of the physical characteristics of the human face or face parts. More subtle biometric-based approaches exploit other or additional physiological and behavioural characteristics, including electroencephalogram, cardiogram, or gesture signals (Mei and Weihong 2018, Schetinin et al. 2018).

Over three decades of research and development in face recognition allowed to achieve high performance in many applications; nevertheless, the recognition methods are still affected by factors such as illumination, facial expression, and poses as described in (Scherhag et al. 2017, Uglov et al. 2008, Xie et al. 2017). These are effects of the general ML problem of OOD, when training data do not fully represent test data.

Efficient attempts to find solutions to these problems have been made with ML and their DL branch methods and techniques, especially with those which extract hierarchical and semantic structures hidden in images (Frintrop 2010, Mei and Weihong 2018).

The early FR and FER ML models used the Bag of (Engineered) Features (BOF) methodology, which generates a compact representation of the large-scale image features in which relations between the features are discarded. The representation may be seen either as a vector in the linear feature space or as a feature frequency histogram. The basis for the feature space, or the list of buckets for a histogram, also called terms vocabulary or visual word vocabulary, is constructed from the features extracted from the training set. Features extracted from the test set are associated with the nearest vocabulary terms, and a novel image is represented as a linear combination of the vocabulary terms and the frequency of their occurrence (O'Hara and Draper 2011).

BOF's popularity was built on its simplicity and compactness of the spatially order-less collections of the quantized local image descriptors, which makes BOF implementations algorithmically and computationally lightweight (O'Hara and Draper 2011). BOF has been adopted for computer vision and pattern recognition and, in particular, for face recognition studies (Chen et al. 2017, Passalis and Tefas 2017, Wu et al. 2011), and FER (Kalsum et al. 2018).

However, with the hardware (GPU) development, BOF models became specific niche ones, with DL models taking priority, especially Convolutional Neural Networks. The CNN architecture was introduced in the late 80's (LeCun et al. 1989), but became the "mainstream" architecture for image recognition in general and FR in particular when the AlexNet model (Krizhevsky et al. 2012) won the "ImageNet Large Scale Visual Recognition Challenge" (Alom et al. 2018, Deng et al. 2009) competition. A quickly growing family of CNN models suitable for FR, with deeper architectures, such as VGG (Simonyan and Zisserman 2014, Yu et al. 2016), and wider architectures, such as GoogLeNet and ResNet (Ren et al. 2016, Szegedy et al. 2015), followed the suite. The latter architectures used Directed Acyclic Graph (DAG) architecture to create cascade cells of parallel convolution layers of different parameters to allow their self-organisation (McNeely-White et al. 2020, Szegedy et al. 2016).

More recently, Transformer architecture, based on attention to local features (Luong et al. 2015, Vaswani et al. 2017), initially introduced into Natural Language Processing (NLP), was adopted for image recognition (Dosovitskiy et al. 2020). Also, hybrid CNN and Transformer architectures were proposed (Carreira et al. 2022, Han et al. 2021, Jaegle et al. 2021, Liu et al. 2021).

Despite their higher than CNN's computational demand, Transformers gained their popularity in FR partially to the hardware progress. The progress also brought fully-connected deep ANNs (or Multi-Layer Perceptrons (MLP)) back to FR attention also (Liu et al. 2021, Tolstikhin et al. 2021) Still, CNN is considered a viable architecture for the next decade (Liu et al. 2022).

The early works on FER, also used the engineered feature BOF methods (Berretti et al. 2010, Shan et al. 2009, Whitehill and Omlin 2006). CNNs were later employed to solve the facial expression recognition problem (Kim et al. 2016, Liu et al.2015, Lopes et al. 2017, Mollahosseini et al. 2016, Ng et al. 2015).

Difficulties in the visual expression-only analysis of emotion expression led researchers to consider multi-modal biometric analysis using such information as pulse rate, ECG and EEG (Alarcao and Fonseca 2017, Raheel et al. 2020). Being a generic feature extraction ANN architecture, the CNN models were used for the classification of such hybrid as well (Hammad et al. 2021).

However, BOF algorithms and DL models are poorly generalised in the OOD conditions, especially for FER tasks. Generative AI, a significant part of which is represented by the Transformer-based models, on the one hand, is not directly related to the FR and FER tasks, but, on the other hand, can be used for the training data set augmentation by generating variety synthetic images, hypothetically filling the gaps of original data sets, and reducing the OOD problem.

*2.2 Transformer architecture*

Transformer architecture for ANN, especially for the encoder-decoder (or just encoder) models (Bengesi et al. 2024), gained tremendous popularity with the publication (Vaswani et al. (2017) in which previous works on the attention mechanisms (Bahdanau et al. 2014, Gehring et al. 2016, Luong et al. 2015) were compiled and repackaged into a multi-head model which was dubbed as "Transformer", and applied to the Natural Language Processing (NLP), and in particular Machine Translation (MT) task.

The attention mechanisms, particularly the dot-product proximity mechanism of Vaswani's Transformer, help to solve the bottleneck's inadequate dimensionality problem of auto-encoders. The too-narrow bottleneck could lose important parameters, while too-wide would ineffectively use computational resources for processing low-useful parameters.

For NLP tasks, such parameters could be sequences of words or, rather, their embeddings into the contextual token (feature) space. In other application domains, such as image (Dosovitskiy et al. 2020), video (Liu et al. 2022), time series processing (Verma and Berger (2021), or even graphs (Min et al. 2022), Transformer architecture can also be beneficial, finding important for the task-related parameters, such as images, pixels, temporal signals, or relations.

*2.3 Transformers' problems*

However, despite their popularity, Transformer architectures exhibit a number of problems of various kinds; some of them are effectively solved in a practical sense, and some are open discussion topics, such as their poor generalization under the Out-of-Distribution (OOD) conditions (Yadlowsky et al. 2023), catastrophic loss of dimensionality, i.e. degradation to a rank-1 matrix over multiple layers (Dong et al. 2021), loss of plasticity and forgetting (Pelosin et al. 2022, Shang et al. 2023), to be fair, that latter one is a problem in general for Deep Learning (DL) architectures.

On top of these high-level issues, more technical problems were dealt with in various variations of Transformer architectures. The so-called quadratic complexity of Transformers or dot-product attention comes from the transposed "key" and "query" matrices multiplication, which, given high-dimensionality input, occupies a large (quadratic input dimensionality) amount of memory (Phuong and Hutter 2022), which is the main problem and also strains computation resources on matrix multiplication. For NLP tasks, such large matrices are rather exotic for very large text attention windows of tens of thousand tokens (which became less exotic with the introduction of the Large Language Models (LLM)), but for Computer Vision (CV) tasks, even modest image sizes in the $200 \times 200 - 300 \times 300$ range, cause a problem.

A patchy, so-called "visual words", architecture of the vision transformers, which offered global linear complexity over insignificant local quadratic complexity of the small patches in (Dosovitskiy et al. 2020), and hierarchical sequence of transformer layer, which would create "visual sentences" (Han et al. 2021). These architectures have been followed by a number of various modifications and borrowings back into NLP, primarily in the patch arrangement direction - sparse, overlapping, variable size, and so on (Bertsch et al. 2023, Yu et al. 2023, Zaheer et al. 2021).

Decoder parts, when reconstructing large outputs in CV tasks or LLM applications, may also face large matrix multiplication constraints. The NLP's response to such a challenge was a so-called Masked Language Model (MLM) (Besag 1975, Salazar et al. 2019) when only a portion in the range/target text is revealed for the model in training at a time. Similarly to the patchy "visual words" borrowing into vision Transformers, patchy output revealing of the range/target image was proposed to deal with the decoder large matrix problem (Carreira et al. 2022, Jaegle et al. 2021).

Another approach for dealing with resource constraints of the dot-product attention, is to use another, additive softmax attention mechanism, proposed already in (Bahdanau et al. 2014), but overshadowed by more popular Vaswani's type Transformers. Which, the dot-product-based

Transformers, also suffer from instability during warm-up training, which is usually controlled by the following normalization layers. Such an instability was proposed to be dealt better, by including normalization inside Transformers (Xiong et al. 2020). Yet another improvement vector, targeting the persistent memory integration into Transformer architecture, proposes elements of Recurring Neural Networks (RNN) by introducing external memory that would hold the Transformer out data at one moment of time, and then feed it back into the Transformer's input at the following time step (Bulatov et al. 2022, Wu et al.2022).

*2.4 Dimensionality inflation problem of the context features in auto-encoder models*

The auto-encoder or encoder-decoder ANN architecture intends to extract important features of the input by mapping it into a limited dimensionality feature space of a bottleneck layer, by reconstructing the original input as a "label". Various mechanisms can be used to avoid this direct and inverse mapping to become identity mapping. For example, probabilistically mixing in a random noise, as in Variational Auto Encoder architecture, which development lead to the popular "diffusion" models.

The problem of such architectures is the dimensionality of the bottleneck, which, if overextended, leads to excessive ANN size on the decoder size. On the other hand, if the bottleneck is too squeezed, important features may be lost. Even for less resource-hungry tasks like image processing, Natural Language Processing (NLP), until recently, was facing that problem.

Traditional NLP tokenizing techniques included the preprocessing stage, on which "stop-words" are removed, remaining words are stemmed and lemmatized (converted to canonical dictionary form), and the Bag of Words (BoW), or Bag of Features (BoF) in general setting, algorithm is used to map lemmatized words into a linear vector space, spanned on the most frequent and important words dictionary basis (O'Hara S and Draper BA 2011, Struhl 2015). The whole sentence or a larger text is represented as a linear sum of all token vectors (or so-called "embeddings") (Zhang et al. 2010). Such an approach is resource usage effective but does not count in the sentence or larger text structure. For example, such sentences as: "A dog bites a man", "A man bites a dog", and "Dogs bite men" would be represented by the same embedding.

To introduce implicit elements of the linguistic structures, modern NLP models frequently use context tokenizers (Taylor 1953) of the BERT-like family (Devlin et al. 2018) that use the attention mechanism of the Transformer architecture to build such a context. A simple illustration of the BoW and BERT embedding differences would be the former creating "DOG", "BITE", "MAN", and the latter – "nullDOGbite", "dogBITEman", "biteMANnull", "nullMANbite", "manBITEdog", "biteDOGnull". That solves the BoW's structure blindness problem but greatly increases the dimensionality of the embedding space, which is the starting point of LLMs' high computational demands and size.

To keep with the human reader's attention span and produce a coherent flow of text, LLMs have to use long context windows for MLM training of thousands of words. The brute force use of the whole continuous windows is computationally problematic; therefore, another technique of extracting the most valuable and influential context words on the predicted word gave birth to computationally tractable but still huge LLMs - Attention mechanism (Bahdanau et al. 2014, Gehring et al. 2016, Liu et al. 2022) and its Transformer implementation (Vaswani et al. 2017). In such an approach of "attention", learnable matrices are used to compute cosine or Euclidean distances between the word relevance to the projected prediction over the context window sliding, and the most consistent contributor over time is kept and used, in such a way, reducing computational demand.

We experiment with a similar approach applied to the synthetic image generation via encoder-decoder ANN, radically reducing the bottleneck size to the number of the being controlled parameters. For example, in application to emotion generation - to 8 neurons in the final bottleneck layer, associated to 6 "basic" emotion expressions, neutral expression, and a closed-eye expression.

## 3. Proposed solution

For the Generative AI model, we utilize the encoder-decoder architecture that uses a Cosine Distance Batch Primary Components (BPC) variation of the Transformers' dot-product attention mechanism to perform an implicit feature importance assignment on the encoder side.

An optional additional and more traditional variation of the Cosine Primary Element (CPE) Transformer attention mechanism, intended to select relevant data in the training and test batches, was preliminarily experimented with on a simple MNIST data set, producing promising results. However, on a more diverse and complex data set of facial expressions BookClub, CPE performed worse, additionally consuming limited computational resources and was excluded from the final architecture.

A traditional visual transformer approach is used to address the quadratic complexity of the dot-product family of attention mechanisms. This approach divides the input image into patches, applies dot-product attention separately to each patch, and then combines the results. The following fully connected (FC) layer decreases the output dimensionality before passing information to the next BPC layer.

The meta parameter of the desired emotion expression is supplied together with input image information and then, with residual connections, is appended to each encoder layer. The first Visual Transformer BPC layer mixes meta-parameters into each attention patch, but in the following BPC-FC pair, only FC layer mixes the residual meta-parameters in.

A bottleneck selector of the important features, or rather a cascade of the gradual reduction of the feature dimensionality, was used to guide the decoder's image generation based on the limited number of features of interest.

The decoder side consists of the Learning ReLU (Rectified Linear Unit) activation functions specific to each neuron in the layer. The size of the layer is set according to ANNs composed on the Kolmogorov-Arnold Representation theorem.

We propose modifications of the canonical families of the Transformer attention heads (Vaswani et al. 2017) and Rectified Linear Unit (ReLU) activation functions. Improvements to the former family are aimed at implementing attention not between embeddings in the feature space, but rather attention between features (or dimensions of the features), effectively performing the feature selection in the Lipschitz sense, by projecting them into relevant subspaces.

*3.1 Cosine and Batch Transformers*

Vaswani-type Transformer attention head, or dot-product head, has an arbitrary or optional scaling factor common for all dimensions and embeddings or mini-batch vectors (for example, proportional to the $W$ matrix dimension, etc.), Equation 1.

(1)
$$k = W_k x + w_{k0}, \quad q = W_q x + w_{q0}, \quad v = W_v x + w_{v0}$$

$$B = q^T k / \sqrt{dk}$$
$$y = v \, softmax(B^T)$$

where $x$ is a layer's input, $W_k x$, $W_q x$, $W_v x$ are linear transformations. In this notation, softmax is applied in a "vertical" direction. It should be noted that because vectors $k$ and $q$, in case of the batch training, become matrices $K$ and $Q$, hence $B$ are matrices in the general case.

Such an arbitrary scaling requires using normalisation layers downstream. However, dot-product attention, naturally, as a part of the cosine distance function, can include implicit dimension-wise normalisation Equation 2. The $V$-transformation is optional in the attention head and, for simplification, is not shown - it could be done outside of the attention head or could be omitted entirely.

(2)
$$k = W_k x + w_{k0}, \, q = W_q x + w_{q0}$$
$$q2 = \sum q \otimes q, \, k2 = \sum k \otimes k$$
$$D = \sqrt{q2^T k2}$$
$$B = q^T k \otimes \frac{1}{D}$$
$$y = x \, softmax(B^T)$$

where $q2$ is a sum in the "vertical" vector dimension and is a horizontal vector (not scalar) for the batch training, as well as $D$ is a matrix, also as $B$, which is calculated as a Hadamard division, and $\otimes$ is element-wise multiplication, or Hadamard product.

The Batch Primary Components (BPC) Transformer is similar to vision Transformers (Dosovitskiy et al. 2020) in terms that it attends not to the nearby embeddings but instead to features or embedding projections to individual dimensions Equation 3; therefore, $B$ is a square matrix of the input vector dimensions for Batch Transformer heads, while for the Vaswani's-type Transformer heads - also square matrix, but mini-batch or sequence dimensions.

(3)
$$k = W_k x + w_{k0}, \, q = W_q x + w_{q0}$$
$$q2 = \sum q^T \otimes q^T, \, k2 = \sum k^T \otimes k^T$$
$$D = \sqrt{q2 \, k2^T}$$
$$B = q \, k^T \otimes \frac{1}{D}$$
$$y = x^T \, softmax(B)$$

The term "Batch" is used here as an analogy to Batch Normalisation layers, which use another dimension for normalisation than Layer Normalisation. The "Primary Component" (PC) term is used because the Batch attention mechanism scales up or down individual dimensions of the input value, in contrast to the Vaswani-like "Primary Element" (PE) attention mechanism, scaling whole vectors in the input batch, Equation 2.

In the case of the vision transformers, not only the PC approach, Equation 4 is used (to find the most common or relevant pixels in the images, instead of the most relevant images in the batch), but also an input space $\mathcal{X}$ is mapped to a quotient space of patch subspaces $\mathcal{X}_1, \mathcal{X}_i, \mathcal{X}_n$, and then, each subspace is transformed using the Transformer's attention mapping, and range spaces joined into the output space $\mathcal{Y}$ by the direct sum $\oplus$:

(4)
$$\pi: X \subset R^m \mapsto \cup\{i \in n\}(X_i \subset R^{m/n})$$
$$y: X_i \subset R^{m/n} \mapsto Y_i \subset R^{m/n}$$
$$z = y_1 \oplus \cdots y_i \oplus \cdots y_n$$

where $n$ is a number of attention patches.

Also, tensors Key $k$ and Query $q$ may not be the only results of the linear transformation. They can be made non-linear by adding activation functions (Cheng et al. 2023). In this research, both BPC and PE attention head variations are modified to use Hyperbolic Tangent non-linearity, Equation 5.

(5)
$$k = tanh(W_k x + w_{k0})$$
$$q = tanh(W_q x + w_{q0})$$

*3.2 Plasticity-learning neuron-wise activation units*

Plasticity-learning neuron-wise activation units, for example, Learning Rectified Linear Unit (LrReLU), instead of using constant scalar slop multiplier $a$ as if in Equation 6, use learnable vector $A$, Equation 7.

(6)
$$y = a \times x, z = relu(y): z_i = max(0, y_i)$$

(7)
$$y = A \otimes x, z = relu(y): z_i = max(0, y_i)$$

where $x \in \mathcal{X} \subset R^m$ is a layer's input of dimensionality $m$, $y \in \mathcal{Y} \subset R^m$ is the layer's output of the same dimensionality, $y_i, z_i$ are parameters along the dimension $i \in m$.

The Equation 8 can be used to roughly emulate member $\Phi_q$ from the Equation 10 below, while for the member $\phi_{qp}$ emulation, neuron-specific hybrid of the Learning ReLU with fully connected layer may be used:

(8)
$$y = (A \otimes x), z = W\, relu(y): z_i = max(0, y_i)$$
where $W$ is another transformation matrix.

*3.3 Proposed ANN architectures*

A general-purpose BPC Transformer-based encoder consists of a fully connected layer that mixes input information and metadata that describe it in a few laconic terms, either real numeric or categorical expressed numerically in some form. Then, a BPC Transformer follows, Equations 2-3, and a fully connected layer squeezes the Transformer's output into a narrow bottleneck. The following decoder, firstly, applies the Learning ReLU (or other Learning Unit) to the encoder's bottleneck input to further disable secondary dimensions; then, a fully connected dimensionality expansion layer follows with the attached second Learning ReLU, Equations 7-8. Finally, depending on the regression or classification task, a fully connected layer and Regression layer follow. This generic architecture is used for image and other parameters prediction tasks, with metadata being a number and dimensionality of predictors (such as emotion), Figure 1, where $H \times V$ is an input data dimensionality, $L \times D$ - number and dimensionality of meta-parameters, N - output dimensionality, and K is a bottleneck compression coefficient.

**Figure 1.** *Encoder-decoder architecture with Batch Transformer and Learning ReLU layers.*

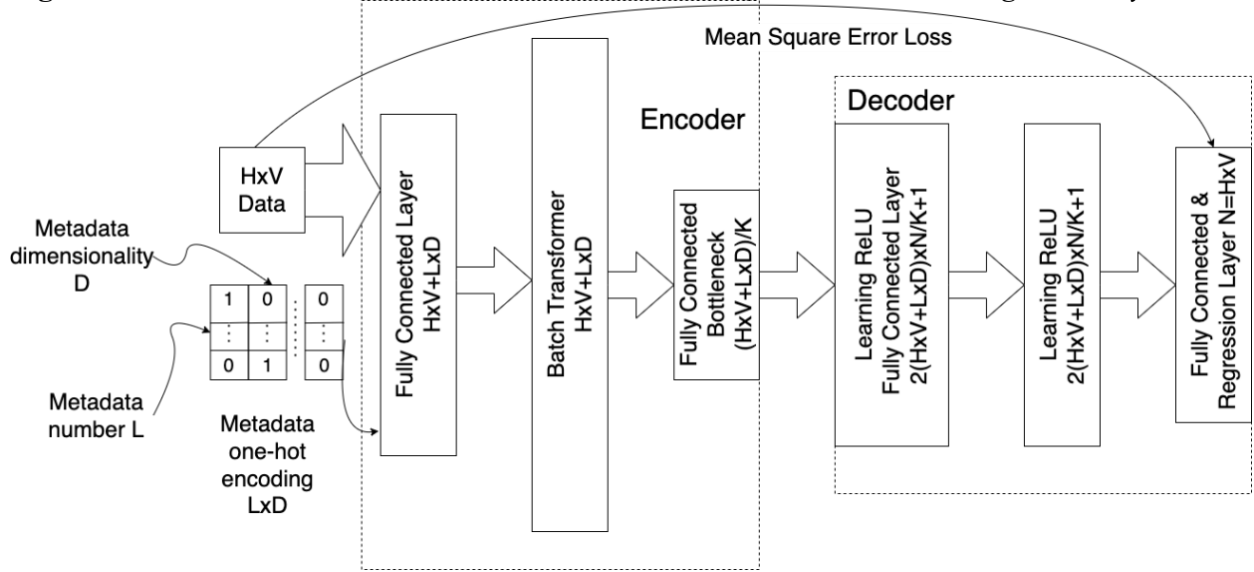

A more specialized architecture that first, applies Cosine Primary Element (CPE) Transformer attention layer to input data, Equations 2, then the similar as described above architecture follows, but with residual or skip connections (previously known as a cascade) (AlFuhaid et al. 1997, He et al. 2015, He R et al. 2020) of time metadata, applied to each consecutive layer, Figure 2.

**Figure 2.** *Encoder-decoder architecture with Cosine and Batch Transformers, residual connections and Learning ReLU layers.*

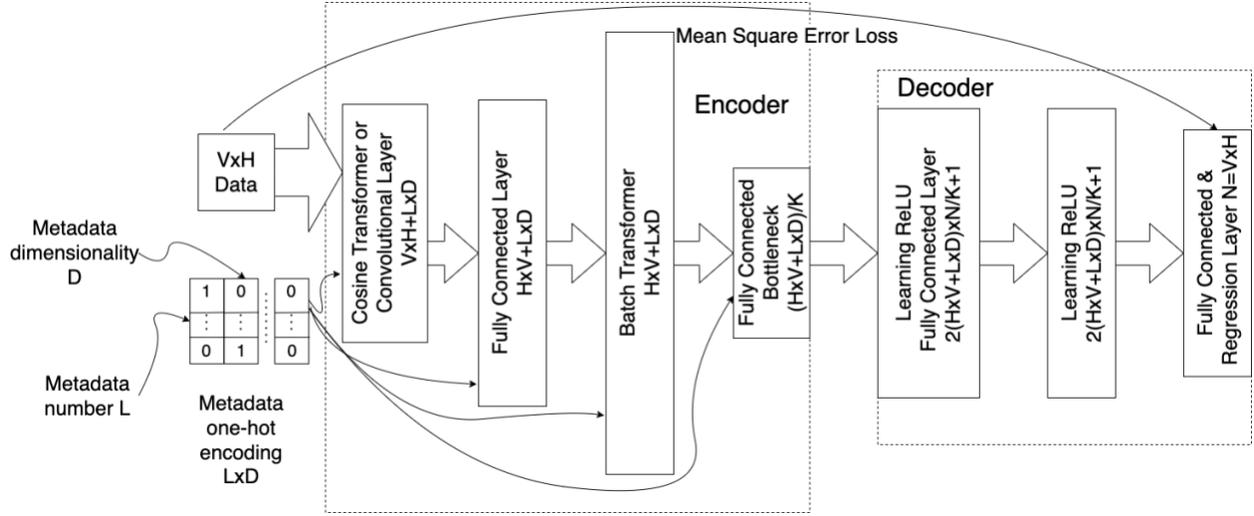

The two hidden non-linearity layers decoder was designed with a number of neurons $m$ and $2m + 1$ on the first and second hidden layer, respectively, where m is a bottleneck dimensionality, Equations 9-10. If we look at the layer transformation of an ANN, as a transformation from a higher dimensional space into the lower dimensional space $f$, $n < m$:

(9)
$$f: \mathcal{X} \subset R^m \mapsto \mathcal{Y} \subset R^n$$

The reason to limit the decoder to two layers was that if one looks at ANN as a Universal Approximation according to the Kolmogorov-Arnold superposition theorem (Kolmogorov 1961), for general emulation of the $f: \mathcal{X} \subset R^m \mapsto \mathcal{Y} \subset R$ process, the 2-layer is a minimal ANN configuration needed (given that activation functions are complex enough):

(10)
$$f(x) = f(x_1, \dots, x_m) = \sum_{q=0}^{2m} \Phi_q \left( \sum_{p=1}^{m} \phi_{qp}(x_p) \right)$$

where $\Phi_q$ and $\phi_{qp}$ are continuous $R \mapsto R$ functions.

The practicality of such ANN, as a Universal Approximator, was disputed in Girosi and Poggio (1989), particularly because of the non-smoothness, hence non-practicality, of the inner $\phi_{qp}$ functions. However, these objections were rebutted in Kurkova (1991). In Pinkus (1999), $\phi_{qp}$ activation functions are even called "pathological". Recently, interest to the Kolmogorov-Arnold architecture-based ANNs has returned (Liu et al. 2024, Vaca-Rubio et al. 2024, Genet and Inzirillo 2024).

Considering $n$ as an ANN's output dimensionality, a $n$-product of Equations 10 was laid into the foundation of the Approximator ANN for Equations 9.

## 4. Data Sets

The proposed simple BPC Transformer-based encoder-decoder architecture, was initially tested as a vision transformer encoder head on the handwritten digits MNIST data set. A more convenient than the original Yann LeCun's repository (2013), a MATLAB format version of the data set was downloaded from (Lulu 2023). For non-continuous training, the original partitioning of the data set was used: 60000 training and 10000 testing images.

After the proof of concept verification on MNIST set, a more challenging BookClub artistic makeup data set was used. The BookClub data set, Figure 3, contains images of number $E = |C| = 21$ subjects. Each subject's data may contain a photo-session series of photos with no makeup, various makeup, and images with other obstacles for facial recognition, such as wigs, glasses, jewellery, face masks, or various headdresses. The data set features 37 photo sessions without makeup or occlusions, 40 makeup sessions, and 17 sessions with occlusions. Each photo session contains circa 168 JPEG images of the $1072 \times 712$ resolution of six basic emotional expressions (sadness, happiness, surprise, fear, anger, disgust), a neutral expression, and the closed eyes photoshoots taken with seven head rotations at three exposure times on the off-white background. The subjects' age varies from their twenties to their sixties. The race of the subjects is predominately Caucasian and some Asian. Gender is approximately evenly split between sessions.

**Figure 3**. *BookClub data set examples.*

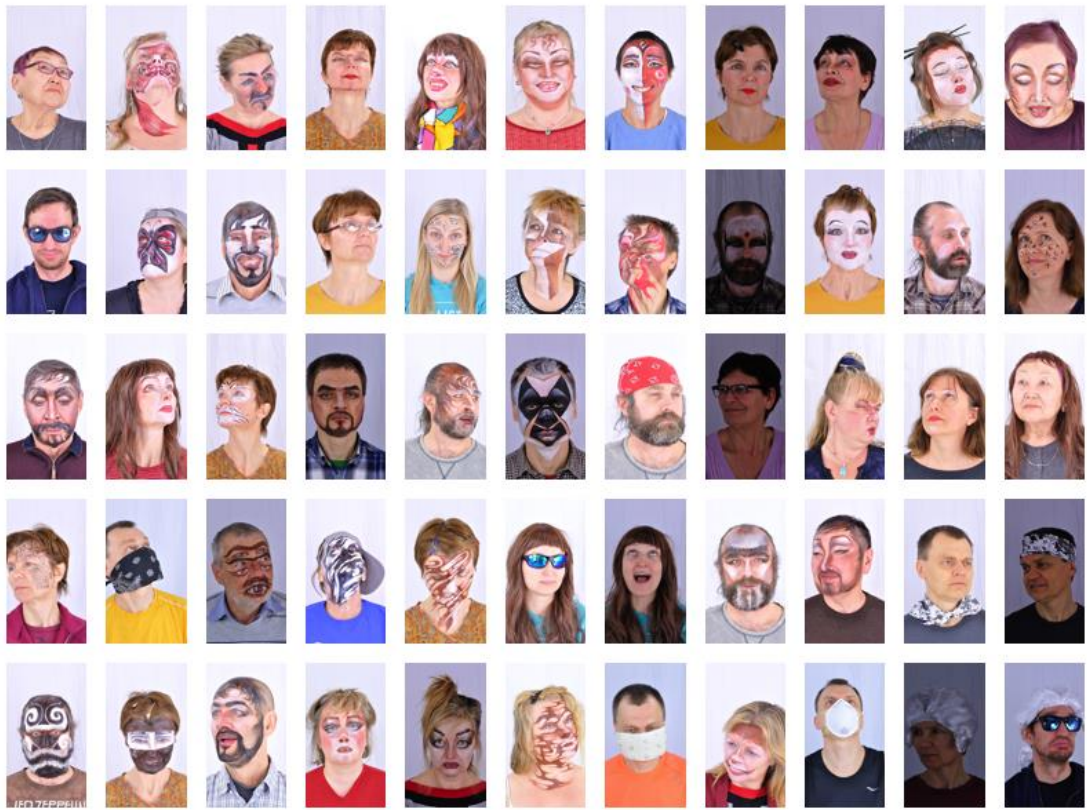

The photos were taken over two months, and several subjects were posed at multiple sessions over several weeks in various clothing with changed hairstyles, downloadable from https://data.mendeley.com/datasets/yfx9h649wz/3. All subjects gave consent to use their anonymous images in public scientific research.

The makeup design and application also varied in the artists' skills, style and heaviness of the pigments and face area covering percentage. Makeup artists of three levels volunteered their work for the project: mature, semi-professional, and professional. As for the pigment materials, professional artistic and theatrical pigments of Mehron, Inc. production were used in experiments. For occlusions, typical everyday items, likely to be found in households and on the streets, were used.

In addition to the practical usefulness in training and verifying ANN against the makeup and occlusions recognition avoidance, when non-makeup-only photo-sessions compound the training set and makeup and occlusion sessions are used for testing, such a data set is suited very well for benchmarking uncertainty estimation for the real-life conditions of the OOD conditions when test data not being well represented by the training data. The wide variety of lighting conditions, head orientations, emotion expressions, age, gender, and race makes this data set an excellent source of data for the aleatoric uncertainty training.

For synthetic image generation experiments, the data set was partitioned into the subject, makeup, and time-centred photo sessions. All images with makeup and occlusions, and a few extra sessions without makeup, were selected for the training sub-set (13980 images), while for the test sub-set, only non-makeup and non-occluded sessions (1653 images) were used.

The number of controlled parameters, such as emotion expression, head rotation, subject, makeup or occlusion type, if used together, create bottleneck size that dictates decoder size putting too much computational requirements for the available GPU. Therefore, such controlled meta-parameters were divided into groups. This contribution describes results for the partition with 8 types of emotion expression, for the front head rotation, and no makeup used. That creates an original training set of 838 images and a test set of 227 images.

## 5. Experiments

The experiments were run on the Linux (Ubuntu 20.04.3 LTS) operating system with QuadroPro RTX 8000 (with 48GB GDDR5 memory), X299 chipset motherboard, 256 GB DDR4 RAM, and i9-10900X CPU. MATLAB 2023b with Deep Learning Toolbox was used as a programming platform. Models were run with 0.001 learning rate. For MNIST model, the minibatch size was 2048 and epoch number - 250, and For BookClub - 64 for minibatch, and 2000 epochs.

For the proof of concept of the proposed BPC Transformer and LrReLU layers, a simple straightforward encoder-decoder architecture was used, Figure 1. The architecture was applied to generating synthetic images using the MNIST data set to estimate the feasibility of the proposed solution visually. Thus, image size is $N \times M = 28 \times 28$, and metadata $L$ was implemented as a one-hot encoding of digits, by setting one of its 10 variables to 1 and leaving others at 0. Randomly selected test images were assigned all 10 metadata values, and were run through the encoder-decoder, generating a mutation table, in which every digit was recreated in all 10 styles.

The color BookClub images were scaled to $100 \times 100 \times 3$, with 3 color channels. For such size, to fit in the GPU, a hierarchical Patchy Vision Transformer encoder organization had to be used, Figure 2. At the first layer of the hierarchical Vision Transformer, $5 \times 5$ patches were used for 3 channels, thus resulting in $m = 3000, n = 75$ $20 \times 20$ BPC Transformers, Equation 3. Then, the output was squeezed by a fully connected layer $K_1 = \frac{1}{3}$ times. The following BPC Transformer and fully connected layers squeezed the data into the final encoder bottleneck $L_b$. The decoder, similarly to the MNIST model, inflated, with coefficient $K_e$, bottleneck data back to the original image size.

The source code is publicly available at https://github.com/Selitskiy/BCGen and https://github.com/Selitskiy/ANNLib.

## 6. Results

The models' run-time was 8 hours for the head-on single Transformer for the MNIST data set, and 2 hours for the vision Patchy BPC Transformer model for BookClub data set.

The mutation table for the randomly selected test digits from MNIST data set is shown in Figure 4. The top row is original images, the following rows are muted by setting rotating metadata one-hot encoding flags for all 10 possible digit modifications.

**Figure 4.** *MNIST digits mutation table. Top row - original images, the next 10 rows - mutated images by metadata encoding rotations.*

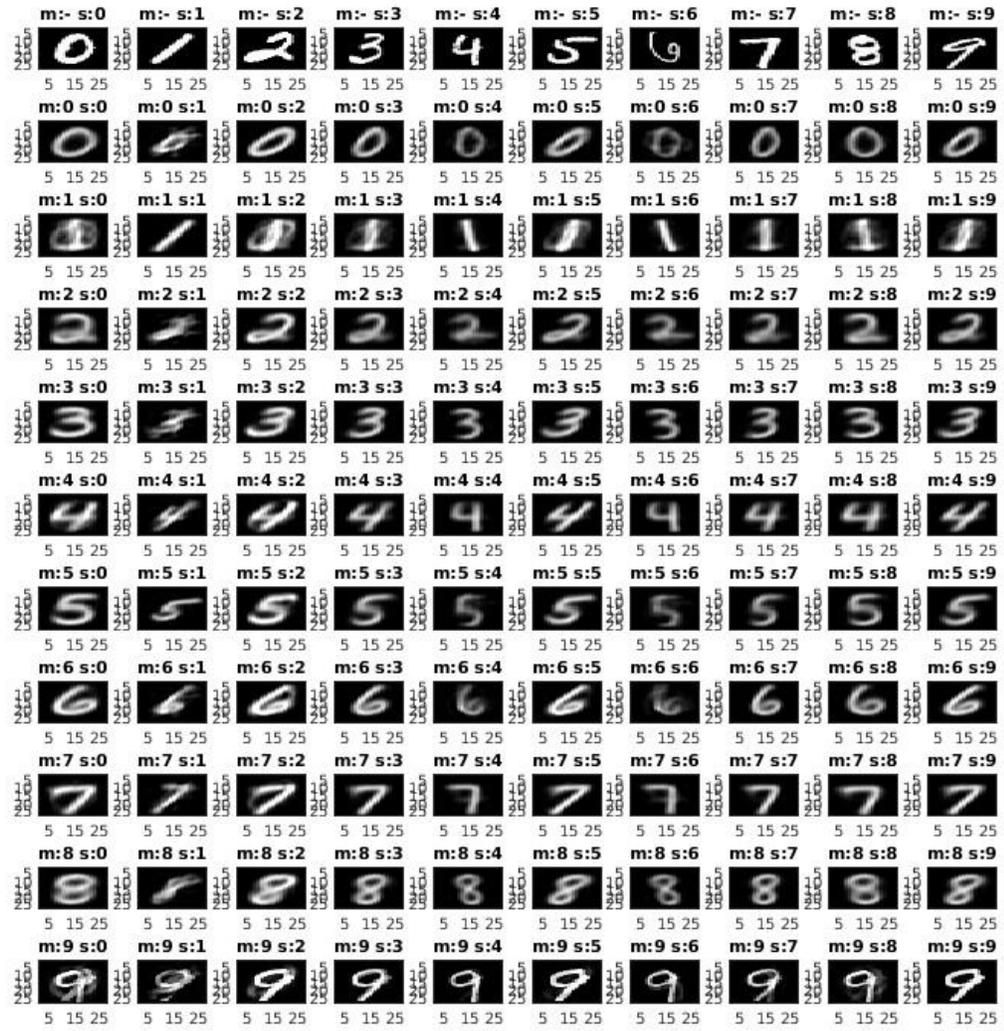

The model for BookClub data set synthetic image generation is illustrated in Figure 5. The top left image is from the test data set, i.e. not seen by the model during training. Because the subject was not the control metadata, a particular subject expressing the generated emotion may be different consistently, or even different for each emotion, Figure 6.

**Figure 5.** *BookClub test image emotion expression mutation. Top left - original image.*

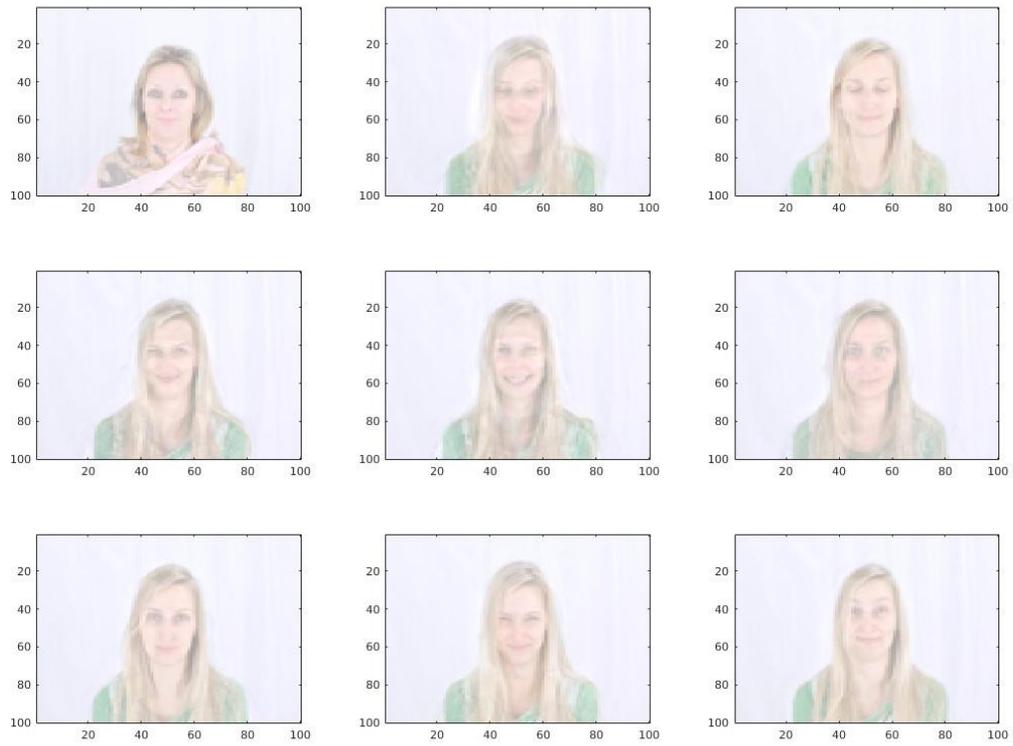

**Figure 6.** BookClub test image emotion expression mutation. Top left - original image. Case of multiple subjects on the generated emotions.

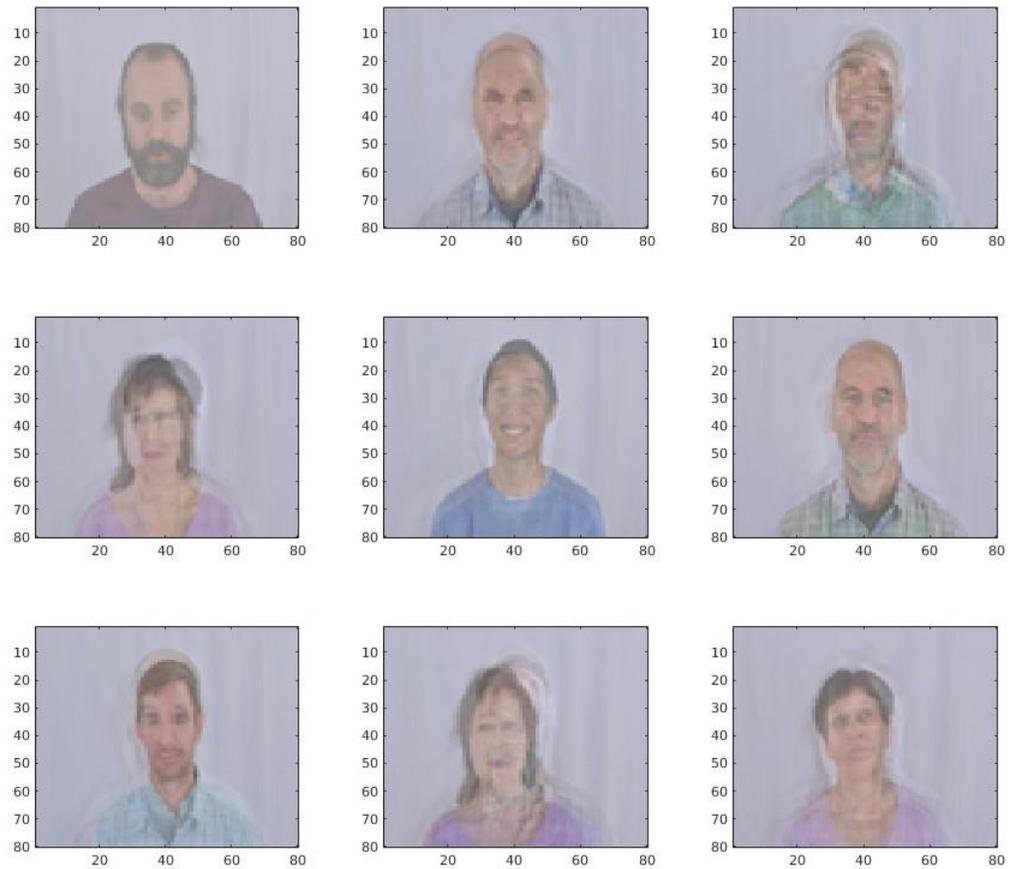

For numerical verification, whether augmenting the original training set would improve FER performance of state-of-the-art (SOTA) CNN models, Inception v.2 model was selected, and pure synthetic training set and combine original + synthetic training sets were experimented with.

Of the 838 original images, synthetically generated emotion expression variants for each original image, made the synthetic training set size 6704, and combined original + synthetic training set - 7542. An example of such emotion mutation for one training set image is shown in Figure 7.

**Figure 7.** *BookClub training image emotion expression mutation. Top left - original image.*

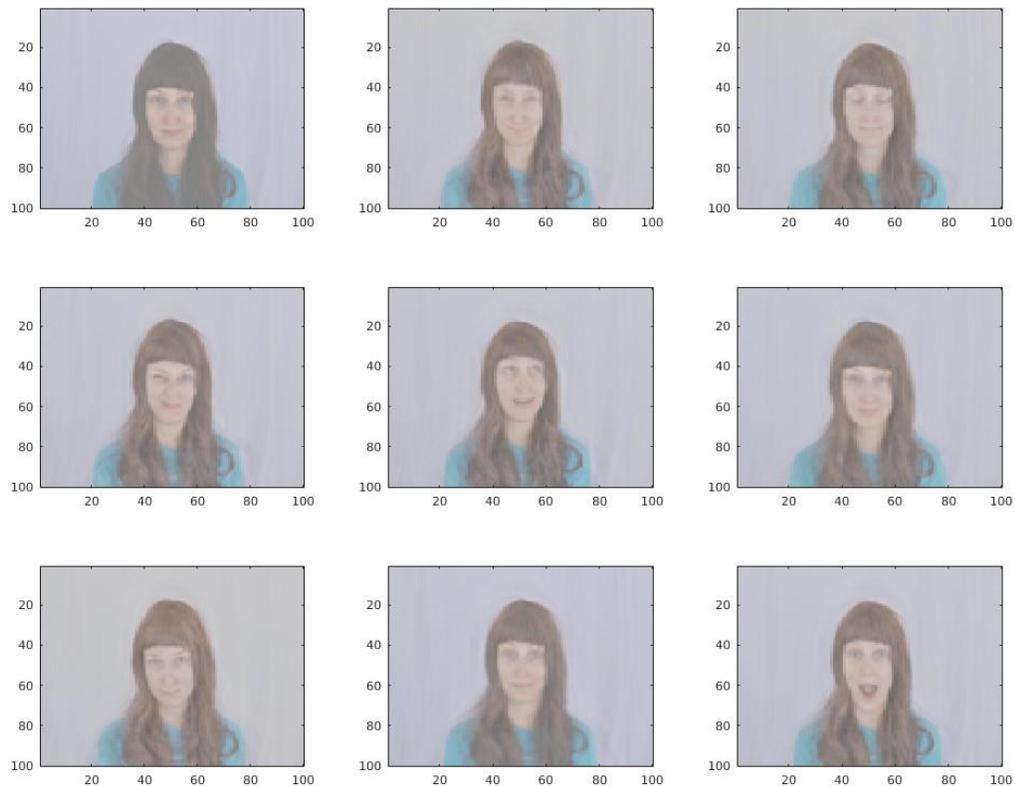

The resulting accuracy of the Inception v.2 model on FER task using original training data set was significantly improved (from 36% to 44% and to 43%) for completely synthetic, and combined original and synthetic data sets, Table.1. For range accuracy estimates, 5-fold variations of the train and test data sets were implemented. For the confidence level (99%) hypothesis testing, the Wilcoxon one-sided signed rank test was applied to the obtained accuracy distributions for the original, synthetic and hybrid trained results of Inception v.3 model.

The results show that at the 99% confidence level p-value is 0.0312, i.e. the probability of the null-hypothesis (that true distributions of synthetic or hybrid trained model results are not better than the originally trained model results) is slightly more than 3%. This means that the hull-hypothesis can be thrown away, and the proposed generative encoder-decoder architecture indeed, statistically significantly improves emotion expression recognition of the Inception v.3 model on the experimented with data set.

**Table 1.** *Accuracy for model Inception v.2 for original, synthetic, and combined training sets.*

| Metric | Original | Synthetic | Combined |
|--------|----------|-----------|----------|
| Accuracy | 0.3602 ± 0.0208 | 0.4392 ± 0.0239 | 0.4329 ± 0.0316 |

## 7. Discussion, Conclusions and Future work

An obvious limitation of the study is a freshly-backed ad-hock models, not yet optimized, and the results presented here are preliminary proof of concept. Another limitation is a visual estimate of the result without rigorous numeric accuracy or efficiency metrics.

Still, it can be seen that the proposed Batch Primary Component Transformer and Learning ReLU-based encoder-decoder architectures work successfully, performing the task of synthetic image generation by mutating the metadata input parameter. Augmenting organic training data sets with synthetically generated data, improves the accuracy of the SOTA CNN model, such as Inception v.2.

Experiments on other data sets are planned for future work, as well as, optimizing models, allowing them to work with larger image sizes, fully controllable by metadata parameters types of image mutation and generation. Also, experiments with other domains as time series and NLP are planned.